\documentclass{article}

\usepackage{PRIMEarxiv}

\usepackage[utf8]{inputenc}
\usepackage[T1]{fontenc}
\usepackage{hyperref}
\usepackage{url}
\usepackage{booktabs}
\usepackage{amsfonts}
\usepackage{nicefrac}
\usepackage{microtype}
\usepackage{graphicx}
\usepackage{multirow}
\usepackage{array}
\usepackage{siunitx}
\usepackage{xcolor}
\usepackage{listings}
\usepackage{longtable}
\usepackage[noabbrev,nameinlink]{cleveref}

\usepackage{amsthm}

\usepackage{subcaption}
\usepackage{caption}
\usepackage{enumitem} 
\usepackage{algorithm}      
\usepackage{algorithmic}    
\usepackage{amssymb}

\definecolor{codegreen}{rgb}{0,0.6,0}
\definecolor{codegray}{rgb}{0.5,0.5,0.5}
\definecolor{codepurple}{rgb}{0.58,0,0.82}
\definecolor{backcolour}{rgb}{0.97,0.97,0.97}

\title{\includegraphics[width=0.99\textwidth]{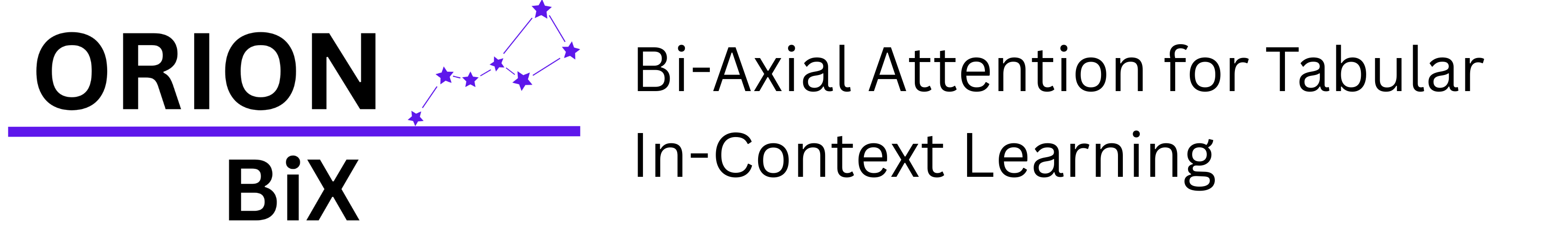}}

\author{
  Mohamed Bouadi, Pratinav Seth\\
  Aditya Tanna, Vinay Kumar Sankarapu \\
  \affiliation{Lexsi Labs, France \& India}
}

\setabstract{%
Tabular data drive most real-world machine learning applications, yet building general-purpose models for them remains difficult. Mixed numeric and categorical fields, weak feature structure, and limited labeled data make scaling and generalization challenging. To this end, we introduce Orion-Bix, a tabular foundation model that combines biaxial attention with meta-learned in-context reasoning for few-shot tabular learning. Its encoder alternates standard, grouped, hierarchical, and relational attention, fusing their outputs through multi-CLS summarization to capture both local and global dependencies efficiently. A label-aware ICL head adapts on the fly and scales to large label spaces via hierarchical decision routing. Meta-trained on synthetically generated, structurally diverse tables with causal priors, Orion-Bix learns transferable inductive biases across heterogeneous data. Delivered as a scikit-learn–compatible foundation model, it outperforms gradient-boosting baselines and remains competitive with state-of-the-art tabular foundation models on public benchmarks, showing that biaxial attention with episodic meta-training enables robust, few-shot-ready tabular learning. The model is publicly available at \url{https://github.com/Lexsi-Labs/Orion-BiX}.
}

\setkeywords{Tabular Foundation Models, Tabular In-Context Learning, AutoML, Benchmarking.}
\runningtitle{Orion-Bix: Bi-Axial Attention for Tabular In-Context Learning}

\begin{document}
\maketitle

\section{Introduction}
Tabular data remain the most common form of real-world data, spanning domains such as healthcare, finance, and scientific research. Despite the remarkable progress of deep learning in natural language processing \cite{liu2023pre,yu2024natural} and computer vision \cite{goldblum2023battle}, gradient boosted trees (GBTs) remain the predominant state-of-the-art (SOTA) for tabular prediction tasks. In other data modalities, foundation models—particularly Large Language Models (LLMs) \cite{DBLP:conf/nips/00020D24, DBLP:conf/nips/LiZLML0HY24}—have significantly advanced the ability to tackle new tasks and few-shot learning. This is largely due to their remarkable in-context learning (ICL) capabilities \cite{DBLP:conf/iclr/Zhang0YOKZY025, DBLP:conf/uss/CarliniTWJHLRBS21}, which enable them to capture patterns directly from prompts without updating their parameters. This success combined with the pervasiveness of tables have spurred interest in tabular foundation models \cite{DBLP:conf/icml/BreugelS24}.

Although LLMs are primarily designed to process natural language, recent efforts have explored fine-tuning them for tabular data tasks \cite{tabllm,tmlr}. These approaches typically rely on table serialization, which is the process of converting table rows into text or sentences suitable for tokenization. For instance, \cite{DBLP:conf/nips/0001PS24} fine-tuned a Llama 3-8B model on a large corpus of serialized tables and demonstrated that this strategy can outperform traditional tree-based models in few-shot scenarios. However, such language model–based approaches face inherent challenges. Their limited context windows restrict the number of serialized examples that can be processed simultaneously (e.g., up to 32 or 64 shots in \cite{DBLP:conf/nips/0001PS24}), and it remains uncertain whether LLMs can reliably interpret and reason over numerical values \cite{DBLP:conf/naacl/ThawaniPIS21}.

Recently, tabular in-context learning has emerged, adapting the ICL paradigm—central to large language models—to tabular data, enabling pretraining across diverse tables and rapid task adaptation without gradient updates. TabPFN \cite{tabpfn} pioneered this approach by meta-training a transformer on synthetic datasets generated via structural causal models. Its encoder--decoder design lets test samples attend to training examples for few-shot prediction, but alternating column- and row-wise attentions make large training sets computationally costly. TabDPT \cite{tabdpt} achieves comparable performance using similarity-based retrieval, though its diffusion process adds overhead. TabPFN-v2 \cite{tabpfn2} extended row-based encoding to datasets exceeding $10{,}000$ samples. In addition, OrionMSP\cite{orionmsp} introduced new architecture that is explicitly designed around feature/sample structure and mixed datatypes using block sparse attention. TabICL \cite{tabicl} proposed a table-native transformer comprising column embeddings, row interactions, and an ICL head. Its SetTransformer-based column encoder and label-aware in-context learner achieve state-of-the-art results, but its row encoder (\texttt{tf\_row}) applies a single homogeneous attention over all features, ignoring local groupings, multi-scale interactions, and structured aggregation. This limitation is pronounced in high-dimensional, heterogeneous tables where features naturally form semantic groups (e.g., demographics, vitals, labs) and dependencies span multiple scales. Additionally, TabICL treats synthetic tables as individual supervised tasks rather than support/query episodes, producing an implicit few-shot signal that may misalign with test-time objectives.

To address these limitations, we introduce \textbf{Orion-Bix}\footnote{The code and model checkpoint are available at https://github.com/Lexsi-Labs/Orion-BiX}, which enhances TabICL with two key improvements: (1) a \emph{biaxial row encoder} replacing \texttt{tf\_row} with complementary attention modes—standard cross-feature, grouped, hierarchical, and relational—aggregated via multiple CLS tokens, explicitly modeling local groups, coarse-scale interactions, and global patterns; and (2) a \emph{meta-learning training regime} that constructs support/query episodes from synthetic tables, optionally using kNN-based support selection, aligning training with few-shot test-time objectives. Orion-Bix preserves TabICL's strengths in column-wise SetTransformer embeddings and label-aware ICL while addressing its row-level and training limitations.
\section{Orion-Bix: Proposed Approach}
\label{sec:orion_bix}
\begin{figure}[p]
    \centering
    \includegraphics[width=1\linewidth,height=0.88\textheight,keepaspectratio]{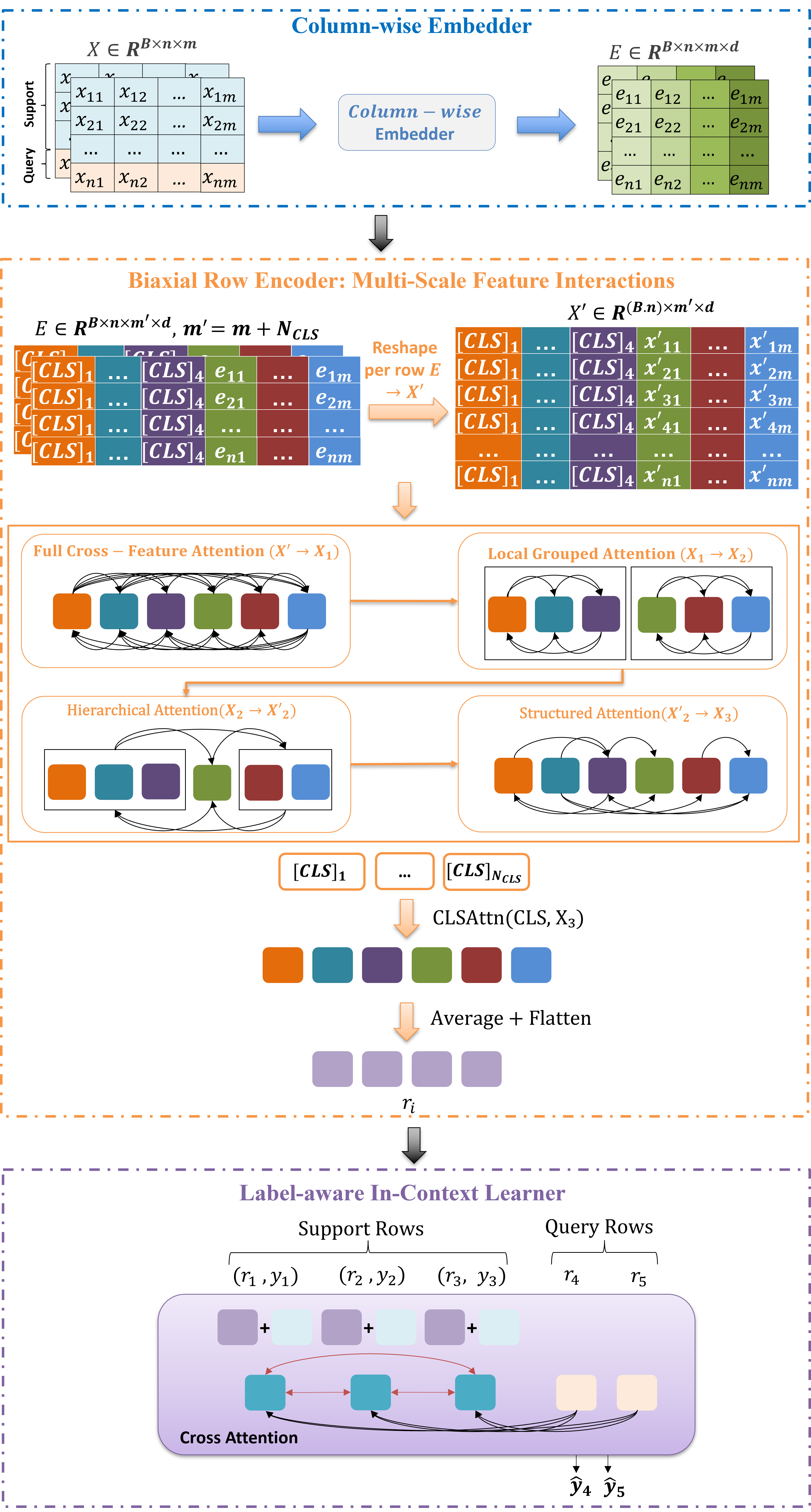}
    \caption{An overview of \textsc{Orion-Bix} architecture. A column-wise SetTransformer-based embedder maps the input table $X \in \mathbb{R}^{B \times n \times m}$ to column embeddings $E \in \mathbb{R}^{B \times n \times m' \times d}$, where $m' = m + N_{\mathrm{CLS}}$ includes reserved CLS slots. The biaxial row encoder reshapes $E$ into per-row sequences $X' \in \mathbb{R}^{(B \cdot n) \times m' \times d}$ and applies a stack of BiAxialAttentionBlocks combining full cross-feature attention  ($X' \rightarrow X_1$), local grouped attention ($X_1 \rightarrow X_2$), hierarchical attention across coarse feature partitions  ($X_2 \rightarrow X'_2$), and structured relational attention ($X'_2 \rightarrow X_3$). A multi-CLS attention layer $\mathrm{CLSAttn}(\mathrm{CLS}, X_3)$ aggregates each row into a multi-aspect representation $R \in \mathbb{R}^{B \times n \times (N_{\mathrm{CLS}} \cdot d)}$. The label-aware ICL adds projected support labels to support row embeddings and uses a masked cross-attention Transformer to predict labels for query rows.}
    \label{fig:archi}
\end{figure}

Orion-Bix retains two core TabICL \cite{tabicl} components: the SetTransformer column encoder and the label-aware in-context learner, while introducing two major changes: (1) a biaxial row encoder that replaces the single-stage \texttt{tf\_row} and provides structured multi-scale feature reasoning, and (2) a meta-learning training regime that constructs explicit support/query episodes from synthetic tables, optionally using kNN-based support selection. An overview of the complete architecture is shown in Figure~\ref{fig:archi}.

\subsection{Problem Setting and Data Representation}

We consider a tabular dataset as a sequence of rows with heterogeneous columns. For each task (dataset), let:
\begin{itemize}
    \item \(X \in \mathbb{R}^{n \times m}\): a table with \(n\) rows and \(m\) features (after preprocessing).
    \item \(y \in \{0,\ldots,C-1\}^{m}\): labels over \(C\) classes.
\end{itemize}
We split rows into a \emph{support set} and a \emph{query set}. The first \(n_{\text{train}}\) rows form the support set, the remaining \(n_{\text{test}} = n - n_{\text{train}}\) rows form the query set. During meta-training and inference, Orion-Bix receives the full table \(X\) and only the support labels
\[
    y_{\text{train}} = y_{1:n_{\text{train}}}.
\]
The goal is to predict labels for the query rows conditioned solely on the support rows and their labels, i.e., in a tabular in-context learning regime.

In practice, Orion-Bix operates on a \emph{batch of tasks}. Inputs have shape
\[
    X \in \mathbb{R}^{B \times n \times m},
\]
where \(B\) is the number of tasks (tables) in the batch. A per-task feature-count tensor
\[
    d \in \mathbb{N}^B
\]
tracks how many of the \(H\) columns are active in each table (the remaining columns are padding).

\subsection{Column-wise Embedding}
\label{sec:col-embed}

We adopt the original TabICL \cite{tabicl} column-wise embedder to map each feature to a \(d\)-dimensional representation per cell. The embedder uses a \emph{Set Transformer}, \(TF_{\text{col}}\), that treats each column as a permutation-invariant set of values across all rows.

\subsubsection{Input and CLS Reservation.}
The column embedder receives \(X \in \mathbb{R}^{B \times n \times m}\). We reserve a fixed number \(N_{\text{CLS}}\) of feature slots (typically \(4\)) at the beginning of the feature dimension to host class tokens at a later stage. These slots are padded with a special ``skip'' value, so they do not influence the subsequent projections.

\subsubsection{Skippable Linear Projection.}
Each scalar value \(x_{b,n,m}\) is passed through a \emph{skippable} linear layer:
\[
    \tilde{x}_{b,n,m} = 
    \begin{cases}
        \text{skip\_value}, & \text{if } x_{b,n,m} = \text{skip\_value},\\
        w_h x_{b,n,m} + b_h, & \text{otherwise},
    \end{cases}
\]
where \(w_h, b_h\) are learnable parameters. This yields an intermediate representation
\(
    \text{src} \in \mathbb{R}^{B \times m' \times n \times D},
\)
with \(m' = m + N_{\text{CLS}}\) and embedding dimension \(D\). The skip logic ensures that missing or padded entries do not pollute the learned representations.

\subsubsection{SetTransformer Over Rows for Each Column.}
Columns are treated as small sets of row values: for each feature \(m\), we consider
\(
    \{\tilde{x}_{b,1,m}, \ldots, \tilde{x}_{b,n,m}\}
\)
and apply a stack of SetTransformer-like attention blocks along the row dimension. The implementation supports two attention types:
\begin{itemize}
    \item \textbf{Linear attention}, with a configurable feature map (e.g., ELU, identity, or hedgehog), enabling \(O(n)\) complexity and stable behavior for long sequences.
    \item \textbf{Induced self-attention}, where a small set of learned inducing points mediate interactions between rows, reducing complexity from \(O(n^2)\) to approximately \(O(n)\).
\end{itemize}

A key design choice is that \emph{support/query leakage is controlled at this stage}. During training, the SetTransformer can be restricted to attend only to support rows; at inference, it can optionally embed support and query jointly depending on a configuration flag.

\subsubsection{Feature-Wise Weight--Bias Parameterization.}
Rather than outputting a single vector per column, the column embedder produces \emph{feature-specific affine parameters}. For each \((b,n,m)\), two linear heads produce:
\[
    W_{b,n,m} \in \mathbb{R}^D, \qquad b_{b,n,m} \in \mathbb{R}^D,
\]
and the final per-entry embedding is
\[
    e_{b,n,m} = x_{b,n,m} \cdot W_{b,n,m} + b_{b,n,m}.
\]
This yields an embedding tensor
\(
    E \in \mathbb{R}^{B \times n \times m' \times D}.
\)
The learned weights and biases capture column-wise statistics, scaling behavior, and type information, and allow the model to normalize and warp distributions in a data-driven way.

\subsection{Biaxial Row Encoder: Multi-Scale Feature Interactions}

The column embeddings \(E \in \mathbb{R}^{B \times n \times m' \times d}\) encode per-cell feature representations. The row encoder aggregates these into row-level summaries suitable for in-context learning. 

The original TabICL row encoder \texttt{tf\_row} applies a single stack of Transformer encoder layers directly over the feature dimension. While effective, this design treats all features as a flat sequence and relies on one homogeneous attention pattern to model: (i) local relations within semantically related feature subsets; (ii) long-range interactions between distant columns, and (iii) global aggregation into a single row representation.

In high-dimensional, heterogeneous tables, this uniform treatment leads to several limitations:
\begin{itemize}
    \item \textbf{Lack of explicit grouping:}
          Tabular features often form natural groups (e.g., demographics, vitals, lab tests), but a single attention layer has no explicit mechanism to focus attention within these groups before reasoning across them.
    \item \textbf{Insufficient multi-scale structure:}
          Capturing both local and global patterns with the same attention pattern can force the model into compromises, especially when some dependencies are short-range (within groups) while others are long-range (across groups).
    \item \textbf{Representation bottleneck:}
          Collapsing a row into a single vector after a flat attention stack can obscure distinct modes of information (e.g., rare but important features vs.\ common global signals).
\end{itemize}

To overcome this, Orion-Bix replaces the flat row encoder with a \emph{biaxial attention module} that applies multiple, specialized feature-space attentions and aggregates their outputs through multi-CLS tokens, enabling structured, multi-scale row representations.

\subsubsection{Input Reshaping.}
Starting from the column embeddings
\(E \in \mathbb{R}^{B \times n \times m' \times d}\),
we reshape to treat each row's features as a sequence:
\[
    E \rightarrow X' \in \mathbb{R}^{(B \cdot n) \times m' \times d},
\]
where each element \(X'_{i} \in \mathbb{R}^{m' \times d}\) represents one row's sequence of feature embeddings.

\subsubsection{Bi-Axial Attention Block.}
Each \(\mathrm{BiAxialAttentionBlock}\) applies four attention modules in sequence over the feature axis:

\begin{enumerate}
    \item \textbf{Standard attention:} full cross-feature self-attention capturing generic dependencies:
        \[
            X_1 = \mathrm{Attn}_{\text{std}}(X_0, X_0, X_0),
        \]
    \item \textbf{Grouped attention:} local attention within \(G\) feature groups of size \(\lfloor m'/G \rfloor\). Within each group, we apply a local attention:
        \[
            X_2 = \mathrm{Attn}_{\text{group}}(X_1),
        \]
    This sharpens \emph{local} interactions between features that naturally cluster.
    \item \textbf{Hierarchical attention:} coarse-scale interactions between two large partitions of the feature set \(X_2 = [X_2^{(1)}; X_2^{(2)}]\):
        \[
            \hat{X}_2^{(1)} = \mathrm{Attn}_{\text{hier}}(X_2^{(1)}, X_2^{(2)}, X_2^{(2)}),
        \]
        \[
            \hat{X}_2^{(2)} = \mathrm{Attn}_{\text{hier}}(X_2^{(2)}, \hat{X}_2^{(1)}, \hat{X}_2^{(1)}).
        \]
    This enforces \emph{coarse-scale} interactions between large blocks of features, useful when long-range dependencies span different column subsets.
    \item \textbf{Relational attention:} a second full self-attention over already structured features:
    \[
        X_3 = \mathrm{Attn}_{\text{rel}}(\hat{X}_2, \hat{X}_2, \hat{X}_2),
    \]
    another full self-attention over features, now operating on representations that already encode both local and hierarchical structure.
\end{enumerate}

\subsubsection{Multi-CLS Aggregation.}
After the four feature-attention passes, each block introduces \(N_{\text{CLS}}\) learnable class tokens:
\[
    \mathrm{CLS} \in \mathbb{R}^{N_{\text{CLS}} \times D}.
\]
For each row (each element of the batch), these CLS tokens are broadcast and used as queries in a dedicated \(\mathrm{CLSAttention}\) module:
\[
    \mathrm{CLS}' = \mathrm{CLSAttn}(\mathrm{CLS}, X_3),
\]
where CLS tokens attend over the feature sequence, followed by a small feed-forward network with residual connections and normalization. The output of one \(\mathrm{BiAxialAttentionBlock}\) is thus a set of CLS summaries per row:
\[
    \mathrm{CLS}' \in \mathbb{R}^{(B \cdot n) \times N_{\text{CLS}} \times D}.
\]

Stacking \(L_{\text{row}}\) such blocks refines these CLS tokens iteratively. Finally, we flatten the CLS dimension:
\[
    R' \in \mathbb{R}^{(B \cdot n) \times (N_{\text{CLS}} \cdot D)},
\]
and reshape back to per-task shape
\[
    R \in \mathbb{R}^{B \times T \times D_{\text{row}}}, \quad D_{\text{row}} = N_{\text{CLS}} \cdot D,
\]
which serves as the input to the in-context learner.

This biaxial design allows Orion-Bix to jointly model:
\begin{itemize}
    \item local interactions within feature groups,
    \item structured cross-group dependencies at a coarse level, and
    \item global relational patterns across all features,
\end{itemize}
while avoiding the information bottleneck of a single pooled vector via multiple CLS tokens.

\subsection{In-Context Learner}
The final module, denoted \(\mathrm{ICLearning}\), performs in-context inference: given row representations and support labels, it predicts labels for the query rows.

\subsubsection{Label Injection.}
Given row encodings \(R \in \mathbb{R}^{B \times n \times D_{\text{row}}}\) and support labels
\(
    y_{\text{train}} \in \{0,\ldots,C-1\}^{B \times n_{\text{train}}}
\),
we embed labels into the same space as rows:
\[
    \ell_{b,n} = \mathrm{LabelEmbed}(y_{\text{train},b,n}) \in \mathbb{R}^{D_{\text{row}}},
\]
where \(\mathrm{LabelEmbed}\) is implemented as a one-hot encoding followed by a linear projection (the \(\mathrm{OneHotAndLinear}\) layer). These label embeddings are \emph{added} to the support row representations:
\[
    R_{b,t} \leftarrow R_{b,t} + \ell_{b,t}, \quad \text{for } t < n_{\text{train}}.
\]
This explicitly conditions the row features on the observed labels while leaving query rows unlabeled.

\subsubsection{ICL Encoder with Split Attention Mask.}
The label-conditioned sequence \(R\) is processed by an encoder (stack of attention blocks) over the row dimension. To respect the support/query structure, we use an integer mask \(n_{\text{train}}\) that implements:
\begin{itemize}
    \item The first \(n_{\text{train}}\) positions (support rows) can attend \emph{only} to each other.
    \item The remaining \(n_{\text{test}}\) positions (query rows) can attend to all support rows but \emph{not} to other query rows.
\end{itemize}
Formally, for a row index \(t\) and key index \(s\),
\[
    \text{mask}(t,s) =
    \begin{cases}
        0, & t < n_{\text{train}},\ s < n_{\text{train}},\\
        0, & t \ge n_{\text{train}},\ s < n_{\text{train}},\\
        -\infty, & \text{otherwise}.
    \end{cases}
\]
This mask is applied inside the attention kernels (both standard and linear attention variants) to enforce the desired information flow. Support representations cannot depend on queries, and each query prediction depends only on support rows and its own features.

\subsubsection{Decoder and Hierarchical Classification.}
The encoder outputs
\(
    Z \in \mathbb{R}^{B \times n \times D_{\text{row}}}.
\)
A small MLP decoder maps each row to class logits:
\[
    \hat{y}_{b,t} = f_{\text{dec}}(Z_{b,t}) \in \mathbb{R}^{C_{\max}},
\]
where \(C_{\max}\) is the maximum number of classes the model is trained to handle natively.

Real tasks may have \(C > C_{\max}\). Orion-Bix transparently handles such cases via a \emph{hierarchical classification tree}:
\begin{itemize}
    \item For a task with \(C\) unique labels, we recursively group the labels into at most \(C_{\max}\) super-classes at each level, forming internal nodes and leaf nodes.
    \item Internal nodes predict group assignments; leaf nodes specialize in fine-grained labels.
    \item Predictions at test time traverse this tree, combining probabilities using the chain rule to produce calibrated probabilities over all \(C\) original classes.
\end{itemize}
When \(C \le C_{\max}\), Orion-Bix uses the flat decoder directly.

\subsection{Meta-Training with Synthetic Episodic Data}

TabICL is trained on synthetic tabular data drawn from a configurable prior, optimizing a global objective that encourages robust in-context behaviour across a wide range of tasks. However, the original training schedule treats each synthetic dataset as a single supervised task, without explicitly structuring the optimization around support/query episodes. This leads to two limitations:
\begin{itemize}
    \item \textbf{Implicit few-shot signal:}
          The model learns to infer from sequence prefixes, but the few-shot structure (support vs.\ query) is only partially enforced by attention masks; it is not the primary unit of optimization.
    \item \textbf{Uncontrolled support selection:}
          Support rows are not explicitly chosen to be informative or diverse with respect to query rows; many updates are driven by redundant or suboptimal supports.
\end{itemize}

Orion-BiX adopts an explicit \emph{meta-learning} perspective on top of a similar synthetic data prior:
\begin{enumerate}
    \item Synthetic tables are generated from a broad prior over feature counts, label spaces, sequence lengths, and distributions, and treated as a pool of potential tasks.
    \item Episode Construction: An \texttt{EpisodeGenerator} converts these tables into many small \emph{episodes}, each defined by a support set, a query set, and a per-task feature-count \(d\). Episodes are formed either by random splits or by kNN-based support selection that explicitly chooses support rows that are both relevant and diverse with respect to queries.
    \item A \texttt{MetaLearningDataset} yields episodes in manageable chunks, and the \texttt{MetaLearningTrainer} processes thousands of episodes per update via micro-batching, gradient accumulation, mixed precision, and (optionally) distributed data parallelism.
\end{enumerate}

This explicit meta-learning formulation brings several advantages over the original TabICL training:
\begin{itemize}
    \item \textbf{Stronger alignment with the test-time objective:}
          The model is always optimized in the exact regime in which it will be evaluated: given a small support set and a query set, infer query labels purely from in-context information.
    \item \textbf{Better use of synthetic diversity:}
          Each synthetic table can yield many distinct episodes with different support/query splits.
          This increases the effective number of tasks seen during training and exposes the model to a broader spectrum of few-shot situations.
    \item \textbf{Support-set quality control:}
          kNN-based episode construction selects support examples that are both close to and diverse for the query set, reducing the proportion of updates driven by uninformative or redundant supports.
    \item \textbf{Stability and scalability:}
          Micro-batching, gradient accumulation, and AMP allow Orion-Bix to train on large episode counts without exhaustively increasing memory usage, while DDP and checkpoint management make multi-GPU training robust.
\end{itemize}

In summary, Orion-Bix keeps the core strengths of TabICL \cite{tabicl}---column-wise SetTransformer embeddings and a label-aware in-context Transformer---but introduces a biaxial row encoder that better matches the structure of tabular feature spaces and a meta-learning training regime that more directly optimizes for few-shot in-context performance.

\subsection{Inference Pipeline and Practical Interface}

For practical deployment on real-world tabular datasets, Orion-BiX is wrapped in a scikit-learn--compatible classifier that automates preprocessing and uses an ensemble of transformed views.

\subsubsection{Preprocessing and Feature Engineering.}
Given an input table \(X\), the wrapper:
\begin{itemize}
    \item detects numerical and categorical features and converts all columns to numeric form;
    \item imputes missing numerical values (e.g., with medians) and handles categorical missingness consistently;
    \item optionally applies one of several normalization schemes (none, power transform, quantile normalization, robust scaling);
    \item clips outliers beyond a configurable \(z\)-score threshold;
    \item applies feature shuffling strategies (none, circular shift, random, Latin patterns) to build diverse column orders.
\end{itemize}

\subsubsection{Ensemble of Transformed Views.}
The preprocessor constructs multiple transformed ``views'' of the dataset, each corresponding to a choice of normalization method and feature permutation. For each view:
\begin{enumerate}
    \item a support/query split is formed (e.g., using a subset of training points as the support set),
    \item the transformed table is passed through the Orion-BiX model, which produces logits or probabilities for the query rows.
\end{enumerate}
Across views:
\begin{itemize}
    \item logits are re-aligned to correct for any class shifts induced by permutations,
    \item predictions are averaged over ensemble members,
    \item an optional temperature-scaled softmax converts logits to probabilities.
\end{itemize}
This ensemble scheme improves stability and robustness, especially for datasets with skewed distributions or strong feature-order effects.

\paragraph{Summary.}
In summary, Orion-BiX combines:
\begin{itemize}
    \item a \textbf{column embedder} that learns distribution-aware feature embeddings,
    \item a \textbf{biaxial row encoder} that models feature interactions at several structural scales and compresses them into multiple CLS summaries, and
    \item a \textbf{label-aware in-context learner} that uses masked attention and hierarchical classification to handle few-shot tasks with arbitrary label spaces,
\end{itemize}
all trained via episodic meta-learning on diverse synthetic tables and exposed through an inference pipeline that integrates seamlessly with standard tabular workflows.
\section{Experimental Evaluation}
\label{sec:evaluation}

We evaluate Orion-Bix against TabICL and other baselines to highlight the benefits of biaxial attention and meta-learning. We run all methods via \textsc{TabTune}~\cite{tabtune}, which standardizes preprocessing and training/evaluation protocols for reproducibility. Our experiments focus on three aspects: (i) domain-specific performance on datasets with natural feature structure, (ii) support set quality to assess robustness, and (iii) few-shot learning curves across varying support sizes. These analyses target the improvements motivating Orion-Bix: structured multi-scale feature interactions and enhanced few-shot adaptation.

\subsection{Evaluation Setup}

\subsubsection{Datasets.} 
We construct domain-specific evaluation suites by grouping datasets from public benchmarks (e.g., \textbf{TALENT}~\cite{talent} and \textbf{OpenML-CC18}~\cite{openmlcc18}) according to application domain, allowing assessment in contexts where feature structure and heterogeneity are most relevant. The domains are: Medical Finance. These domains exhibit the high-dimensional, structured, and multi-scale features that Orion-Bix is designed to handle.

\subsubsection{Evaluation Metrics.} 
For each dataset, we report overall classification accuracy (ACC), class-weighted F1 to account for imbalance, and mean rank across datasets within each domain based on accuracy. All models use official train/test splits unless noted. In-context learning models (TabICL, Orion-Bix, TabPFN, TabDPT, Mitra, ContextTab) follow the standard protocol: support examples are provided, and predictions on the test set are made without gradient updates.

\subsection{Domain-Specific Performance}

\begin{table}[hbpt]
  \centering
  \scriptsize
  \caption{Domain-specific performance for Medical, Finance and Energy datasets from the benchmark suites. Formatting: Bold = 1st place; \underline{underlined}= 2nd place within each group.}
  \label{tab:domains_mf}
  \begin{tabular}{l ccc ccc ccc}
    \toprule
    \multirow{2}{*}{Models} & \multicolumn{3}{c}{Medical} & \multicolumn{3}{c}{Finance} \\
    \cmidrule(lr){2-4} \cmidrule(lr){5-7} \cmidrule(lr){8-10}
     & Rank & ACC & F1 & Rank & ACC & F1 \\
    \midrule
    XGBoost & 6.32 & 0.7834 & 0.7669 & 6.62 & 0.7958 & 0.7885 \\
    RandomForest & 6.38 & 0.7779 & 0.7752 & 7.32 & 0.8052 & 0.8001 \\
    CatBoost & 6.36 & 0.7784 & 0.7594 & \underline{5.82} &  0.8117 & \underline{0.8015} \\
    LightGBM & 5.32 & 0.7949 & 0.7614 & 6.17 & 0.8095 & 0.7974  \\
    TabICL & 5.54 & 0.7819 & 0.7696 & 6.60 & \underline{0.8125} & 0.7942\\
    \textbf{\underline{Orion-Bix}} & \textbf{\underline{4.10}} & \underline{0.7893} & \underline{0.7759} & \textbf{\underline{5.39}} & \textbf{\underline{0.8206}} & \textbf{\underline{0.8125}} \\
    TabPFN & \underline{5.04} & \textbf{\underline{0.7984}} & \textbf{\underline{0.7857}} & 7.17 & 0.8094 & 0.7919  \\
    Mitra & 10.77 & 0.3935 & 0.2863 & 13.67 & 0.5340 & 0.4250 \\
    ContextTab & 8.66 & 0.6681 & 0.6129 & 11.25 & 0.7430 & 0.6834 \\
    TabDPT & 6.86 & 0.7764 & 0.7641 & 8.00 & 0.8080 & 0.7960 \\
    \bottomrule
  \end{tabular}
\end{table}

Table~\ref{tab:domains_mf} reports results on Medical, Finance, and Energy domains. Orion-Bix achieves the best mean rank in Medical (4.10) and Finance (5.39), outperforming TabICL and showing competitive performance against gradient-boosted baselines.

\begin{itemize}
    \item \textbf{Medical.} Orion-Bix attains rank 4.10 (vs.\ TabICL 5.54), accuracy 0.7893, and F1 0.7759. While TabPFN slightly exceeds accuracy (0.7984), Orion-Bix’s superior ranking indicates more consistent performance across diverse datasets. The gains reflect the benefit of biaxial attention on grouped and hierarchical features typical in medical data.

    \item \textbf{Finance.} Orion-Bix leads with rank 5.39 (vs.\ TabICL 6.60), accuracy 0.8206 (vs.\ 0.8125), and F1 0.8125 (vs.\ 0.7942). Finance datasets feature strong hierarchies and multi-scale dependencies, where biaxial attention better captures structured relationships, notably improving F1 by +1.83 points over TabICL.\looseness=-1
\end{itemize}

These results highlight two design benefits of Orion-Bix:
\begin{enumerate}
    \item \textbf{Biaxial attention for structured features}: Gains are largest in domains with natural feature groups and hierarchies, confirming that modeling local (grouped), coarse-scale (hierarchical), and global (relational) interactions improves representations over a single attention stack.
    \item \textbf{Multi-CLS aggregation}: Maintaining multiple CLS tokens preserves distinct feature aspects, capturing both rare patterns and common global signals, leading to more robust predictions.
\end{enumerate}

\begin{figure}[ht]
  \centering
  \begin{subfigure}[b]{0.49\linewidth}
    \centering
    \includegraphics[width=\linewidth]{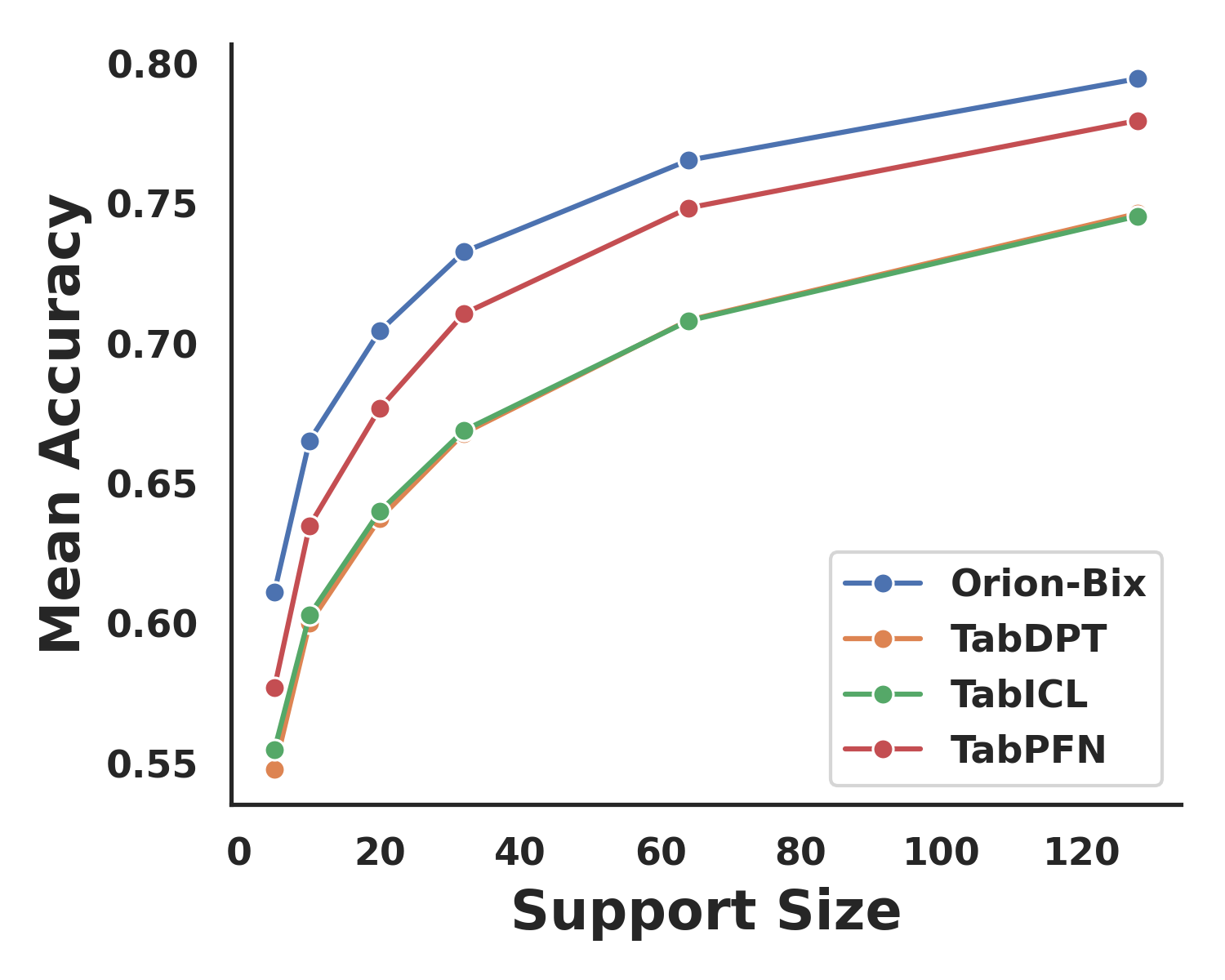}
    \caption{Accuracy vs Support Size on TALENT Benchmark.}
    \label{fig:acc_talent}
  \end{subfigure}
  \begin{subfigure}[b]{0.49\linewidth}
    \centering
    \includegraphics[width=\linewidth]{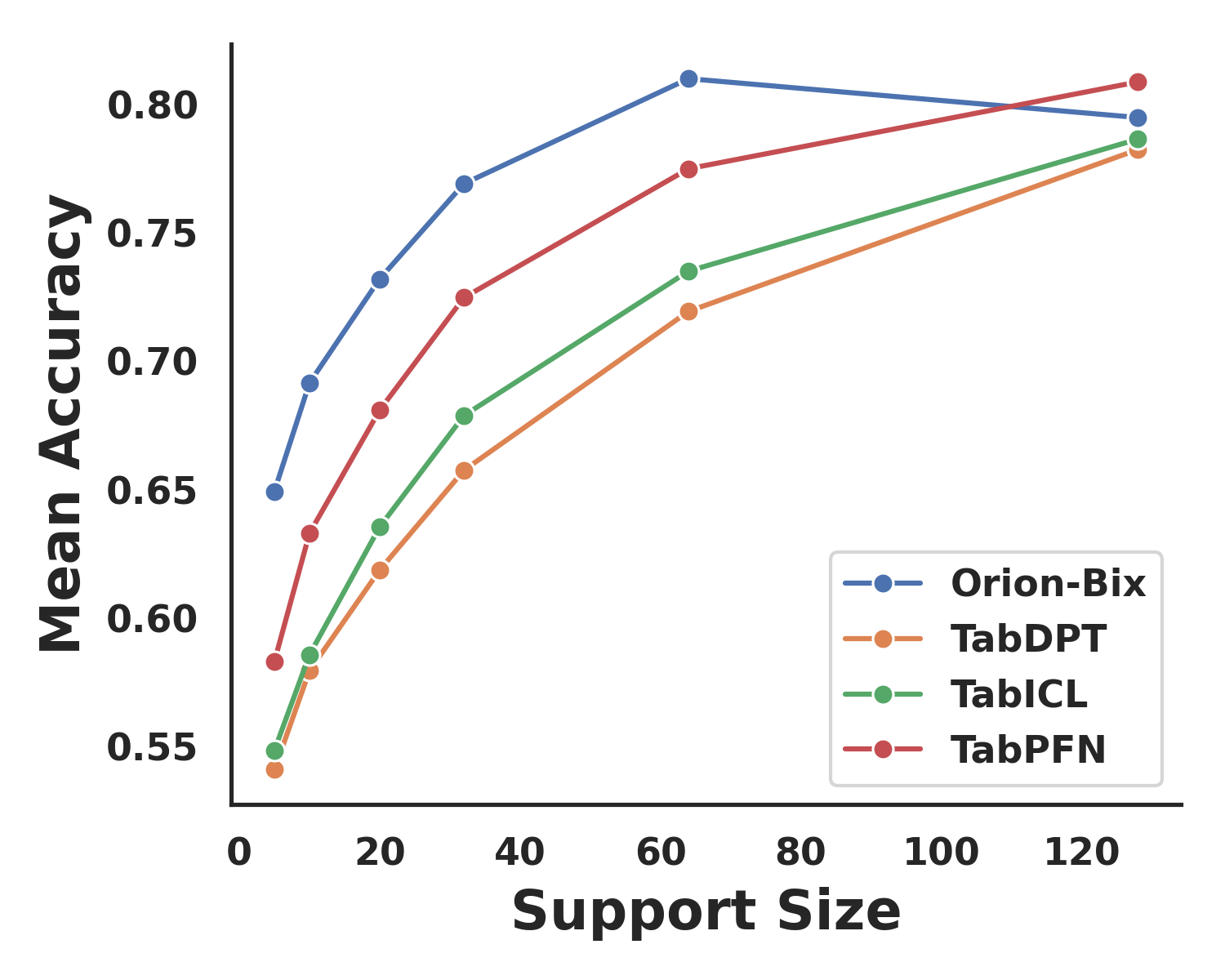}
    \caption{Accuracy vs Support Size on OpenML-CC18 Benchmark.}
    \label{fig:acc_openml}
  \end{subfigure}
  \caption{Accuracy of Orion-Bix and TabICL across different support sizes in few-shot experiments.}
  \label{fig:fewshots}
\end{figure}

\subsection{Few-Shot Performance}

We assess performance as a function of support set size, \\ \(k \in \{5, 10, 20, 32, 64, 128\}\), to test whether Orion-Bix's meta-learning benefits are most pronounced in the very few-shot regime.

\paragraph{Protocol.} For each dataset and support size \(k\), \(k\) examples are sampled from the training set (all classes represented) to form the support set. Models are evaluated on the full test set, repeated across multiple seeds, and results are averaged. Accuracy is then plotted against support size.

\paragraph{Results.}
Figure~\ref{fig:fewshots} shows:
\begin{itemize}
    \item Orion-Bix attains the best accuracy for very low shot sizes (\(k \le 32\)), outperforming all three baselines by 2–4 points, indicating meta-training on diverse episodes improves data efficiency.
    \item Accuracy increases monotonically with \(k\) for all models; gains are largest from \(k=5\) to \(k=32\), while improvements beyond \(k=64\) are small, indicating diminishing returns from additional support examples.
    \item As \(k\) grows, the gap narrows: on TALENT, Orion-Bix stays clearly best across all \(k\), while on OpenML-CC18 TabPFN continues improving and slightly surpasses Orion-Bix at \(k=128\). In both cases, TabPFN and Orion-Bix remain well ahead of TabICL and TabDPT. Overall, Orion-Bix is strongest in the few-shot regime.
\end{itemize}

\subsection{Support Set Quality Analysis}
\begin{figure}[H]
    \centering
    \includegraphics[width=0.8\linewidth]{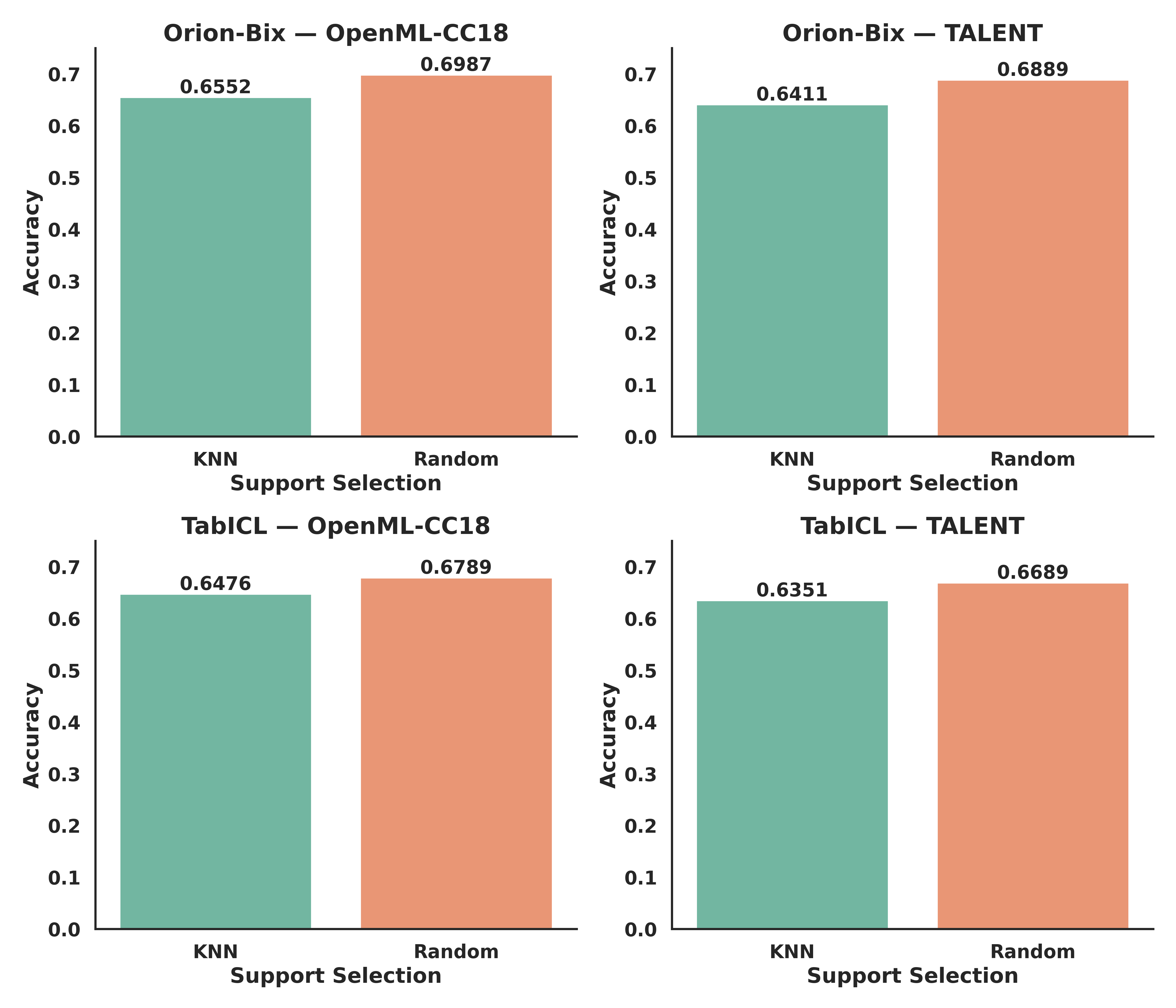}
    \caption{Accuracy of Orion-Bix and TabICL on OpenML and TALENT benchmarks. Bars indicate support selection strategies, averaged over datasets for each model.}
    \label{fig:support_quality}
\end{figure}

We study how support set selection affects few-shot performance by comparing random vs.\ diverse (kNN-based) sampling \emph{at inference}. This tests whether Orion-Bix's meta-learning, which constructs diverse support/query episodes during pre-training, improves robustness to support quality.

\paragraph{Protocol.}
Each test dataset is split 80/20 into train/test sets. A support set of size \(k=32\) is drawn using: (i) \textbf{Uniform sampling}, ensuring all classes are represented, and (ii) \textbf{Diverse (kNN-based)}, selecting examples close to test queries and maximally diverse. Models are evaluated with standard ICL on the same support/query splits, repeated across multiple seeds for statistical significance.

\paragraph{Motivation.} This protocol measures \emph{inference-time robustness}: diverse selection simulates a scenario where practitioners can actively choose informative support examples, while random selection represents default, arbitrary support sampling.

\paragraph{Results.}
Figure~\ref{fig:support_quality} shows that: (1) Uniform sampling slightly outperforms kNN-based selection (by 3–5 points), indicating few-shot performance is not highly sensitive to support curation; (2) Orion-Bix consistently exceeds TabICL under both schemes, with relative ordering unchanged, suggesting meta-training improves overall in-context generalization rather than depending on a specific support selection heuristic.

\section{Conclusion}
\label{sec:conclusion}

Orion-Bix advances tabular in-context learning by strengthening how a model represents \emph{row-level} structure and how it is \emph{trained} for few-shot use. Specifically, it replaces TabICL's flat row encoder with a biaxial row encoder that explicitly mixes (i) full cross-feature attention, (ii) grouped local attention, (iii) hierarchical cross-partition attention, and (iv) relational global attention, and then aggregates these complementary interactions through multi-CLS summaries to preserve multiple aspects of each row representation. This design directly targets common failure modes of tabular transformers in high-dimensional tables, where features naturally form groups and dependencies span multiple scales, while avoiding the information bottleneck of collapsing everything into a single pooled vector.

In addition, Orion-Bix reframes training around explicit episodic meta-learning. Rather than treating each synthetic dataset as one supervised task, training constructs many support/query episodes from synthetic tables via an episode generator (optionally with kNN-based support selection), aligning optimization with the exact inference setting used at test time. This episode-based regime is designed to improve data efficiency and robustness in the few-shot regime by exposing the model to a broader variety of support/query configurations and by optimizing directly for in-context generalization.

Empirically, Orion-Bix shows consistent gains in domains where structured feature interactions matter. On domain-grouped benchmarks, it improves mean rank over TabICL in domains such as Medical and Finance while remaining competitive with strong tree-based baselines. Its advantages are most pronounced in the low-shot setting

\subsection{Limitations}
\label{subsec:limitations}

Despite these improvements, several limitations remain:

\begin{enumerate}
    \item \textbf{Evaluation scope is primarily classification-centric:}
    The experiments emphasize classification metrics such as accuracy and class-weighted F1. Further validation on regression, ranking, calibration, and decision-focused metrics would better characterize real deployment performance.

    \item \textbf{Support-selection benefits are limited:}
    Although the training procedure allows kNN-based episode construction, inference-time results indicate that uniform sampling can slightly outperform kNN-based selection, suggesting curated support selection is not a guaranteed win and may be sensitive to dataset geometry or distance metrics.

    \item \textbf{Potential computational overhead and tuning complexity:}
    The biaxial encoder uses multiple attention passes (standard, grouped, hierarchical, relational) and multi-CLS aggregation. While this improves representational structure, it may increase implementation complexity and require careful tuning (e.g., number of groups, number of CLS tokens, episode size) to balance accuracy and efficiency across datasets and hardware.

    \item \textbf{Large label spaces rely on hierarchical routing:}
    Orion-Bix handles tasks where the number of classes exceeds the native decoder capacity via hierarchical label grouping and tree-based prediction. This introduces another layer where suboptimal grouping could affect accuracy or calibration, especially for fine-grained or highly imbalanced label sets.
\end{enumerate}


\bibliographystyle{plain}
\bibliography{refrences}

\end{document}